\documentclass[runningheads]{llncs}
\usepackage{graphicx}
\usepackage{amsmath,amssymb} 
\usepackage{color}
\usepackage{multirow}
\usepackage{url}
\usepackage{cite}
\usepackage{subfigure}
\usepackage[utf8x]{inputenc}
\usepackage[width=122mm,left=12mm,paperwidth=146mm,height=193mm,top=12mm,paperheight=217mm]{geometry}
\begin{document}
\pagestyle{headings}
\mainmatter
\def\ECCV18SubNumber{492}  

\title{A Deep Learning based Framework to Detect and Recognize Humans using Contactless Palmprints in the Wild} 

\titlerunning{Technical Report No. COMP-K-24}

\authorrunning{Technical Report No. COMP-K-24}

\author{Yang Liu, Ajay Kumar\newline
Department of Computing\newline
The Hong Kong Polytechnic University, Hong Kong 
}
\institute{Technical Report No. COMP-K-24, March 2018}

\maketitle

\begin{abstract}
Contactless and online palmprint identfication offers improved user convenience, hygiene, user-security and is highly desirable in a range of applications. This technical report details an accurate and generalizable deep learning-based framework to detect and recognize humans using contactless palmprint images in the wild. Our network is based on fully convolutional network that generates deeply learned residual features. We design a soft-shifted triplet loss function to more effectively learn discriminative palmprint features. Online palmprint identification also requires a contactless palm detector, which is adapted and trained from faster-R-CNN architecture, to detect palmprint region under varying backgrounds. Our reproducible experimental results on publicly available contactless palmprint databases suggest that the proposed framework consistently outperforms several classical and state-of-the-art palmprint recognition methods. More importantly, the model presented in this report offers superior generalization capability, unlike other popular methods in the literature, as it does not essentially require database-specific parameter tuning, which is another key advantage over other methods in the literature.
\keywords{Biometrics, Contactless Palmprint Matching, Contactless Palmprint Detection, Personal Identification, Deep Learning}
\end{abstract}

\section{Introduction}

Automated personal identification using palmprint images has been widely studied and employed for a range of law-enforcement and e-security applications. However \textit{contactless} palmprint identification is relatively new area of research and offers more attractive solution for the deployments as it can address serous concerns relating to the hygiene while offering higher convenience and user security. In addition, the contactless palmprint imaging also enables deformation free acquisition of palmprint features, or the ground truth information, which can enable higher matching accuracy than those using contact-based imaging.

Despite strong motivation and desire to develop contactless palmprint identification solutions, there are several challenges that needs to be addressed by researchers. Firstly, the palmprint matching accuracy degrades relatively for the contactless images as such palmprint images generally presents higher imaging variations. Therefore more advanced matching techniques needs to be developed to improve the matching accuracy from the contactless palmprint images. Secondly, the detection of contactless palmprint images (region of interest) from the presented hands is quite challenging as the background during such imaging is expected to be dynamic or less stable. Available research on contactless palmprint images addresses such challenges by acquiring contactless palmprint images with fixed background that can enable key point detection using pixel-wise operators to segment the palmprint images. Deep learning capabilities offer enormous potential to address these two challenges and are considered in this report.

In recent years, deep learning has emerged as the dominant approach for a range of computer vision related problems and has delivered state-of-the-art performance for object detection \cite{Ren2017Faster, Girshick2015Fast}, face recognition \cite{Schroff2015FaceNet, he2016deep}, iris recognition \cite{zhao2017towards} and image classification. However, unlike for the face recognition, there is almost nil attention to incorporate remarkable capabilities from the deep learning for palmprint identification and achieve superior performance than popular or state-of-the-art palmprint recognition methods. 

This report proposes a new, deep learning based, contactless palmprint identification framework which not only offers accurate matching capabilities but also exhibits outstanding generalization capabilities on different public databases. With the design of effective residual feature network, our model can enlarge the receptive field \cite{li2018csrnet} for matching contactless palmprint images and learn comprehensive palmprint features which generalizes very well on other databases. We develop a soft-shifted triplet loss function to accommodate frequent contactless palmprint imaging variations and offer meaningful supervision for learning effective palmprint features from limited size of training samples. We also introduce a contactless palm detector to automatically detect palm images, from the presented hands under complex backgrounds, and design of such palm detectors is critical for the success contactless palmprint identification during deployments.

The main contributions in this technical report can be summarized as follows: (a) We develop a new deep learning based contactless palmprint identification framework with high generalization capability for operating on different contactless palmprint databases that can represent diverse deployment scenarios. A new \textit{Soft-Shifted Triplet Loss} (SSTL) function has been developed to successfully address the nature of contactless palmprint patterns for learning comprehensive palm features (please see more details in section 2.3). Our work therefore presents significant advances to bridge the gap between deep learning and contactless palmprint matching techniques available today; (b) Under fair comparison, our approach consistently outperforms several state-of-the-art methods on publicly available contactless palmprint databases. Even under challenging scenario without incorporating any parameter tuning on the target dataset, our model can still achieve superior or competing performance over the state-of-art methods that have been subjected to extensive parameter tuning. This report also demonstrates how the faster-R-CNN \cite{Ren2017Faster} architecture can be adapted to build an online palm detector, which can robustly detect palm images from the presented hands under complex backgrounds. Such advancement is highly desirable, in the current literature, for the success of online and contactless palmprint identification applications.

\subsection{Related Work}

Completely automated matching of contactless palmprint images has received lot of attention and a range of palmprint matchers have been introduced in the literature. Detected or segmented palm images can be characterized by major/minor curved lines and creases that can be observed from low resolution ($\sim 100$ dpi) images and additional flexion ridges \cite{ashbaugh1991palmar} that are observed from high resolution ($\sim 500$ dpi, not focus of this work like for \cite{dai2011multifeature}) images. Therefore a range of texture matching methods have been introduced in the literature \cite{michael2008touch, ribaric2005biometric, morales2011towards, zheng20163d, sunordinal, zhang2017towards}. Encoding palmprint features using the dominant orientation of lines/creases in \cite{kong2004competitive,jia2008palmprint} is one of the most effective method for matching palmprint images. More recent work in matching contactless palmprint images appear in T-PAMI16 \cite{zheng20163d} where an ordinal measurement based descriptor, i.e., difference of normal (DoN), has shown to outperform a range of methods introduced for matching contactless palmprint images using publicly available databases. This approach benefits from the contactless palm image acquisition modeling and introduces specialized masks to encode projective ordinal measurements. Therefore, this method has also been used to ascertain the effectiveness of approach developed in this technical report and serves as a reasonable choice as other methods have not yet shown to offer superior performance than from \cite{zheng20163d} in the best of our knowledge.

Automated detection of palm images, or the region of interest from the hands presented by users, is inherently required for the success of automated palmprint identification systems. Most popular methods for palmprint detection are based on the extraction of key-points representing finger joints and extract a fixed region of interest relative to the orientation and/or the distance \cite{zhang2003online} between the key points. This approach works very well for the contact-based imaging setups but poses a range of problems for contactless palmprint images as its very difficult to robustly detect these key-points under background changes which are inherent during the contactless imaging even with the cooperative users attempting access. Therefore developed contactless palmprint databases \cite{iitd_url,casia_url,polyu_iitd_url} (in public domain) have been acquired using relatively fixed or stable background to primarily address the open problem of detecting palm images under user friendly contactless imaging setup. More work to detect contactless palmprint under contactless imaging is highly desirable and is also considered in our work.

\subsection{Open Problems and Challenges}
Despite promising performance indicated in the literature for matching palmprint images, conventional palmprint descriptors have several limitations. Summary of earlier work presented in \cite{zhang2012comparative} indicate that existing methods offer quite accurate performance but this performance needs to be further improved (especially on large contactless databases \textit{e.g.} \cite{polyu_iitd_url}) to meet expectations for a wide range of deployments. Conventional palmprint descriptors, such as CompCode \cite{kong2004competitive} or DoN \cite{zheng20163d}, RLOC \cite{jia2008palmprint} or Ordinal \cite{sunordinal}, are based on empirical models, which apply hand crafted filters for the generation of features. Therefore these models heavily rely on the parameter selection when incorporated for matching performance for other/different databases or those acquired under different imaging environments. This situation can also be observed from \cite{zheng20163d}, where eight different combination of parameters or 4 different databases are employed by extensive tuning. Commonly employed techniques in the the palmprint literature \cite{zhang2003online,wang2016contactless} for the automated detection of palm images, or the region of interest from the hands presented by users interested to access the system, often fails when the hand images are acquired under complex backgrounds. Such failure can be attributed to the nature of algorithms that relies on the detection of key-point using pixel-based operators that are dependent to differentiate gray-levels from skin and the background.

The deep learning based approaches have potential to address above outlined limitations, since the parameters in deep neural networks are not empirically set instead self-learned from the data and the deep learning architectures are known to offer high generalization capabilities. However, any direct application of such architectures, e.g. \cite{kumar2016identifying}, is expected to deliver limited performance or cannot match performance offered from state-of-art techniques such as those from \cite{zheng20163d}. This is due to the fact that new challenges emerge while incorporating typical deep learning architectures (e.g. CNN) for the palmprint recognition, which can primarily be attributed to the nature of palmprint patterns. Unlike the face, palmprint patterns are known to reveal little structured information or meaningful hierarchies. The palmprint texture is widely considered to be more accurate methods in the literature \cite{zheng20163d,sunordinal,zhang2017towards,kong2004competitive,jia2008palmprint,zhang2012comparative} which mainly employed small sized filters or block based operators to extract palmprint features. Therefore, we can infer that the most discriminative information from palmprint patterns is extracted from the local intensity distributions in region of interest (palm) images rather than from (if any) global features. The CNNs are known to be effective in recovering features from low level to the high level, and from local to global, due to the combination of convolutional and fully connected layers \cite{sun2014deep}. However as outlined earlier, the high level and global features extracted from such networks may not be optimal for the accurate matching of palmprint patterns.

This report attempts to develop a more accurate and robust deep learning based palmprint feature representation framework and makes significant contributions towards fully discovering the potential from the deep learning for the contactless palmprint identification. Such objectives are yet to be pursued in the literature. Different from \cite{he2016deep, zhao2017towards, kumar2016identifying}, this technical report develops a novel deep network and customized loss function, which are highly optimized to extract discriminative palmprint features and has been comparatively evaluated with several state of art methods using multiple public contactless palmprint databases.

\section{Matching Contactless Palmprint Images}

\subsection{Network Architecture}

\begin{figure}
  \setlength{\abovecaptionskip}{1pt}
  \setlength{\belowcaptionskip}{0pt}
  \begin{minipage}[t]{0.5\linewidth}
  \centering
  \includegraphics[width=2.4in]{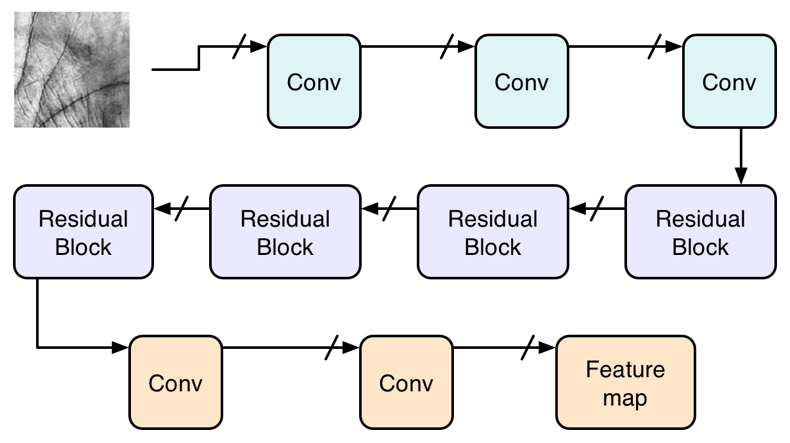}
  \end{minipage}%
  \begin{minipage}[t]{0.5\linewidth}
  \centering
  \includegraphics[width=2in]{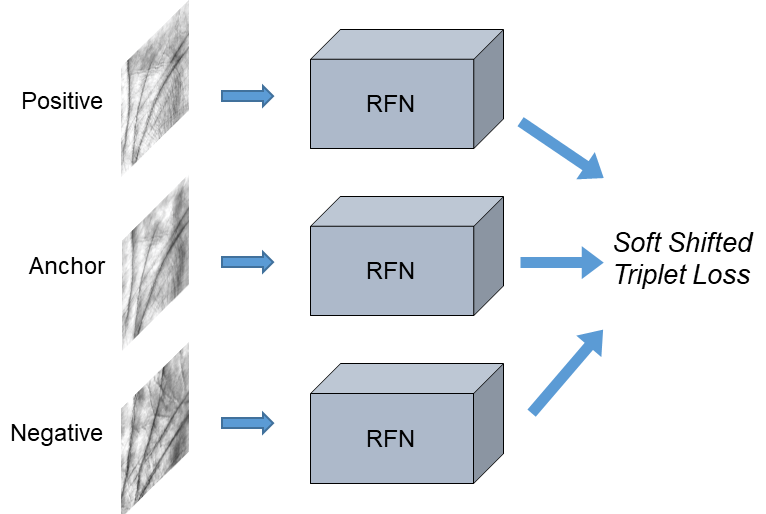}
  \end{minipage}
\caption{\textbf{Left}: Detailed architecture of residual feature network for contactless palmprint matching. The RFN generates a single-channel feature map for each of the input images. The first and the second convolutional layer down-sample the input which results in the feature map that is of one quarter the size of input; \textbf{Right}: Training the RFN using a triplet-based network configuration.}
\label{fig:arch}
\end{figure}

We develop a highly optimized deep learning architecture, referred to as residual feature network (RFN) in this report, for accurately matching contactless palmprint images. Different from the residual network \cite{he2016deep}, RFN does not have fully connected layers which results in pure feature map outputs (Figure \ref{fig:arch}.(a)) that can preserve spatial-correspondences with palmprint images. As illustrated in Figure \ref{fig:arch}.(b), we replace all of the batch normalization layers \cite{ioffe2015batch} with the instance normalization \cite{huang2017arbitrary}. Our key motivation is to enhance the robustness of RFN in learning low/mid/high level features \cite{zeiler2014visualizing} as the contactless palmprint images present significant variations due just due to deformations \cite{wu2015deformed} but also due to the pose and illumination changes \cite{polyu_iitd_url}.
\vspace{-0.5cm}


\subsection{Network Training}
The convolutional kernels of RFN were trained using a triplet network \cite{Schroff2015FaceNet}. As shown in Figure \ref{fig:arch}, this triplet network consists of three identical RFN’s and their weights are kept identical during the training. These RFN’s are inter-connected in parallel to enable the forward and backward propagation of the data and gradients for anchor, positive and negative samples respectively. The triplet loss function in such architecture is expected to help the network learn in generating the feature maps that can reduce the anchor-positive distances while increase the anchor-negative distances. Learning feature maps for the accurate matching of contactless palmprint images requires us to generate the network loss to accommodate frequent intra-class changes in the contactless palmprint images. We therefore soften the matching loss and improve the original loss function to accommodate frequent translational changes in the contactless palmprint images from the same class/subject. This new loss function is referred to as \textit{Soft Shifted Triplet Loss} (SSTL) and is detailed in section 2.3.
\subsection{Soft-Shifted Triplet Loss Function}

The triplet networks \cite{Schroff2015FaceNet} have been conventionally trained using the original loss function which can be written as follows:
\vspace{-0.1cm}
\begin{equation}
  L=\sum_{i}^{N}[\|\mathcal{F}(I_{i}^{a})-\mathcal{F}(I_{i}^{p})\|^2 - \|\mathcal{F}(I_{i}^{\alpha})-\mathcal{F}(I_{i}^{n})\|^2 + m]_{+}
  \vspace{-0.1cm}
\end{equation}
where the function $\mathcal{F}(I)$ represents the embedding of the input image $I$ into a high dimensional feature space, $N$ is the number of triplet samples in a mini-batch, $\mathcal{F}(I_{i}^{a})$, $\mathcal{F}(I_{i}^{p})$ and $\mathcal{F}(I_{i}^{n})$ are the feature representations of anchor, positive and negative image samples in the $i$-th triplet respectively. The symbol $[\circ]_+$ is equivalent to $max(\circ,0)$. $m$ is preset parameter to control the desired distance between anchor-positive and anchor-negative. For simplicity, we denote these three feature maps from the input $I_{i}^{a}$, $I_{i}^{p}$, $I_{i}^{n}$ as $\mathcal{F}_{i}^{a}$, $\mathcal{F}_{i}^{p}$, $\mathcal{F}_{i}^{n}$ respectively.

Accurate matching of contactless palmprint requires us to match the segmented palmprint images, which generally depict high translational changes along the two axes. In order to accommodate such translations, we alter the original triplet loss function and such new loss function is referred to as the \textit{Soft-Shifted Triplet Loss}(SSTL):
\vspace{-0.1cm}
\begin{equation}
  SSTL=\frac{1}{N}\sum_{i=1}^{N}[\mathcal{L}(\mathcal{F}_{i}^{a},\mathcal{F}_{i}^{p})-\mathcal{L}(\mathcal{F}_{i}^{a},\mathcal{F}_{i}^{n})+m]_{+}
  \vspace{-0.1cm}
\end{equation}
where $\mathcal{L}$ represents the \textit{Minimum Shifted Loss}(MSL). Any loss function to train the network should be differentiable along the shift directions and is detailed in the following.

Let us denote the width and height of the feature map from RFN by $W$ and $H$ respectively. We use $W_s$ and $H_s$ to define extent of maximum expected spatial shifts along the horizontal and vertical directions. The \textit{MSL} is defined to accommodate frequent translational shifts in the input or segmented contactless palmprint images, as follows:
\begin{equation}\label{equ:msl}
  \mathcal{L}(\mathcal{F}_1, \mathcal{F}_2)=\min_{-W_s\leq w \leq W_s,-H_s\leq h \leq H_s}{\{D_{w,h}(\mathcal{F}_1, \mathcal{F}_2)\}}
\end{equation}
\begin{equation}\label{equ:dfunc}
  D_{w,h}(\mathcal{F}_1, \mathcal{F}_2)=\frac{1}{|C_{w,h}|}\sum_{(x,y)\in C_{w,h}}(\mathcal{F}_{1}^{(w,h)}[x,y] - \mathcal{F}_{2}[x,y])^2
\end{equation}
\begin{equation}
  \begin{split}
    C_{w,h}=\{(x,y)|\max(w,0)\leq &x \leq \min(W+w, W),\\ 
    \max(h,0)\leq &y \leq \min(H+h,H)\}
  \end{split}
\end{equation}
where $C$ represents the common region between two matched feature maps with valid (non-zero) values for each of the $(w, h)$ combinations while $x$ and $y$ denotes the spatial coordinates. The \textit{MSL} in (\ref{equ:msl}) attempts to compute the minimum distance between the two feature map that can be achieved after translation by $w$ and $h$ pixels along the horizontal and vertical directions respectively. The superscript $(w, h)$ in (\ref{equ:dfunc}) denotes such translational operation on feature map $\mathcal{F}_1$ and the resulting shifted feature map has following spatial correspondence with the original one:
\begin{equation}\label{equ:6}
  \begin{split}
    \mathcal{F}^{(w,h)}[x_w, y_h]&=\left\{
      \begin{array}{lr}
        \mathcal{F}[x,y], &(x, y)\in C_{w,h} \\
        0, &\text{otherwise}
      \end{array}
      \right. \\
      x_w &= (x-w+W) \text{ mod } W \\ 
      y_h &= (y-h+H) \text{ mod } H
  \end{split}
\end{equation}
$x_w$ is obtained by shifting the feature values to the left (horizontal translation) in a step of $w$ and $y_h$ is obtained by shifting the feature values upward (vertical translation) in a step of $h$. As illustrated in (6), the void generated due to the translation of feature map values are automatically assigned as zeros. The training of RFN requires us to compute the gradients (or partial derivatives) of the soft shifted triplet loss, between the anchor-positive and anchor-negative feature maps. The resulting loss is back propagated iteratively during the network training. Let us firstly consider the loss between the feature map from the anchor and its respective positive feature map $\mathcal{F}_{i}^{p}$ and compute its derivative for the one sample pair in the batch:
\begin{equation}\label{equ:7}
  \frac{\partial SSTL}{\partial \mathcal{F}_{i}^{p}}=\left\{
    \begin{array}{lr}
      0, &\text{ if } SSTL = 0 \\ 
      \frac{\partial SSTL}{\partial \mathcal{L}(\mathcal{F}_{i}^{a}, \mathcal{F}_{i}^{p})}\frac{\partial \mathcal{L}(\mathcal{F}_{i}^{a}, \mathcal{F}_{i}^{p})}{\partial \mathcal{F}_{i}^{p}}, &\text{otherwise} 
    \end{array}
    \right.
\end{equation}
Since $\frac{\partial SSTL}{\partial \mathcal{L}(\mathcal{F}_{i}^{a}, \mathcal{F}_{i}^{p})}=\frac{1}{N}$, above equation can be further simplified as 
\begin{equation}\label{equ:8}
  \frac{\partial SSTL}{\partial \mathcal{F}_{i}^{p}}=\left\{
    \begin{array}{lr}
      0, &\text{ if } SSTL = 0 \\ 
      \frac{1}{N}\frac{\partial \mathcal{L}(\mathcal{F}_{i}^{a}, \mathcal{F}_{i}^{p})}{\partial \mathcal{F}_{i}^{p}}, &\text{otherwise} 
    \end{array}
    \right.
\end{equation}
Let us firstly define the shifting offsets for the anchor-positive and anchor-negative image pairs that can meet requirements for \textit{MSL} as follows:
\begin{equation}
  \begin{split}
    (w_{ap}, h_{ap}) &= \mathop{\arg\min}_{-W_s\leq w \leq W_s,-H_s\leq h \leq H_s}{\{D_{w,h}(\mathcal{F}_{i}^{a}, \mathcal{F}_{i}^{p})\}} \\ 
    (w_{an}, h_{an}) &= \mathop{\arg\min}_{-W_s\leq w \leq W_s,-H_s\leq h \leq H_s}{\{D_{w,h}(\mathcal{F}_{i}^{a}, \mathcal{F}_{i}^{n})\}}
  \end{split}
\end{equation}

The gradient of the distance $\mathcal{L}$ in (\ref{equ:8}) can be computed from the following pixel-wise derivatives using (\ref{equ:msl}) and (\ref{equ:dfunc}):
\begin{equation}\label{equ:10}
  \begin{split}
    \frac{\partial \mathcal{L}(\mathcal{F}_{i}^{a}, \mathcal{F}_{i}^{p})}{\partial \mathcal{F}^p_i[x,y]} &= \frac{D_{w_{ap},h_{ap}}(\mathcal{F}_{i}^{a}, \mathcal{F}_{i}^{p})}{\partial \mathcal{F}^p_i[x,y]} \\
    &=\left\{
      \begin{array}{lr}
        0, \text{ if } (x,y) \notin C_{w_{ap}, h_{ap}} \text{ or } SSTL = 0 \\
        \frac{-2(\mathcal{F}^a_i[x_{w_{ap}},y_{h_{ap}}]-\mathcal{F}^p_i[x,y])}{N|C_{w_{ap}, h_{ap}}|}, \text{ otherwise}
      \end{array}
      \right.
  \end{split}
\end{equation}

The partial derivative of SSTL with respect to the positive feature map $\mathcal{F}_i^p$ can be computed as follows:
\begin{equation}
  \frac{\partial SSTL}{\partial \mathcal{F}^p_i}=\left\{
    \begin{array}{lr}
      0, \text{ if } (x,y) \notin C_{w_{ap}, h_{ap}} \text{ or } SSTL = 0 \\
      \frac{-2(\mathcal{F}^a_i[x_{w_{ap}},y_{h_{ap}}]-\mathcal{F}^p_i[x,y])}{N|C_{w_{ap}, h_{ap}}|}, \text{ otherwise}
    \end{array}
    \right.
\end{equation}

We can similarly compute the required partial derivatives with respect to the negative feature map:
\begin{equation}
  \frac{\partial SSTL}{\partial \mathcal{F}^n_i}=\left\{
    \begin{array}{lr}
      0, \text{ if } (x,y) \notin C_{w_{an}, h_{an}} \text{ or } SSTL = 0 \\
      \frac{-2(\mathcal{F}^a_i[x_{w_{an}},y_{h_{an}}]-\mathcal{F}^n_i[x,y])}{N|C_{w_{an}, h_{an}}|}, \text{ otherwise}
    \end{array}
    \right.
\end{equation}
Our final requirement is to compute the partial derivatives for the feature map from anchor. It can be observed from (\ref{equ:msl})-(\ref{equ:6}) that the shifting or translation of the first map towards the left by $w$ pixels and towards the top by $h$ pixels is  equivalent to shifting the second map towards the right by $w$ pixels and towards the bottom by $h$ pixels. We can therefore rewrite (\ref{equ:dfunc}) as follows:
\vspace{-0.1cm}
\begin{equation}\label{equ:13}
  \begin{split}
    D_{w,h}(\mathcal{F}_1,\mathcal{F}_2)&=\frac{1}{|C_{w,h}|}\sum_{(x,y)\in C_{w,h}}(\mathcal{F}^{(w,h)}_1[x,y],\mathcal{F}_2[x,y])^2 \\
    &=\frac{1}{|C_{w,h}|}\sum_{(x,y)\in C_{w,h}}(\mathcal{F}_1[x,y],\mathcal{F}^{(-w,-h)}_2[x,y])^2
  \end{split}
  \vspace{-0.1cm}
\end{equation}
It is now quite straightforward compute the partial derivative for the anchor positive feature map using (\ref{equ:7})-(\ref{equ:10}) and (\ref{equ:13}):
\vspace{-0.1cm}
\begin{equation}
  \frac{\partial SSTL}{\partial \mathcal{F}^a_i[x,y]}=-\frac{\partial SSTL}{\partial \mathcal{F}^p_i[x_{-w_{ap}},y_{-h_{ap}}]}+\frac{\partial SSTL}{\partial \mathcal{F}^n_i[x_{-w_{an}},y_{-h_{an}}]}
\vspace{-0.1cm}
\end{equation}

The rest of the back-propagation process is the same as for common end-to-end convolutional network. Above derivation shows that during the matching of feature maps, from the translated palmprint images, the gradients that only lie in the overlapped regions will be back-propagated. This can allow more accurate matching of feature maps from the contactless palmprint images that are not strictly aligned. The network is trained using \textit{SSTL} while the \textit{MSL} is used during the test or the evaluation phase.
\section{Experiments and Results}
We performed thorough experiments using publicly available databases to ascertain various aspects of the performance from our approach. In the following sections, we detail on the experimental protocols, along with the reproducible results \cite{ExeCodeAjay}, employed for the extensive evaluation of the model proposed in this report.

\begin{figure}
  \setlength{\abovecaptionskip}{0pt}
  \setlength{\belowcaptionskip}{-10pt}
  \begin{minipage}[t]{0.5\linewidth}
  \centering
  \includegraphics[width=2.2in]{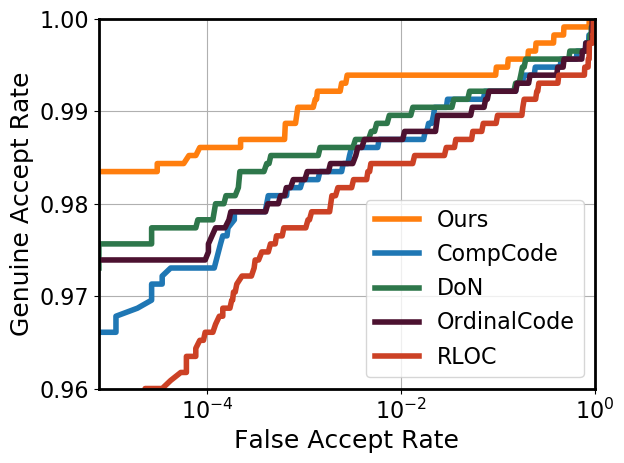}
  \end{minipage}%
  \begin{minipage}[t]{0.5\linewidth}
  \centering
  \includegraphics[width=2.2in]{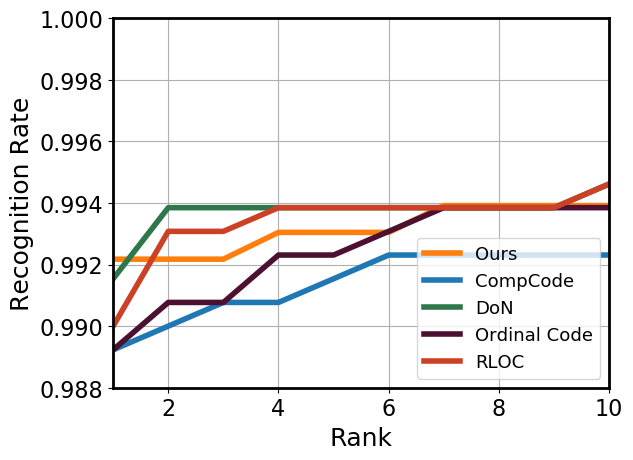}
  \end{minipage}
\caption{The ROC curves (\textbf{Left}) and CMC curves (\textbf{Right}) of different methods from the IITD Right contactless palmprint database.}
\label{fig:iitd}
\end{figure}

Our experiments are firstly organized to ascertain within database performance (\textit{WithinDB}) which uses some part of the database for the training while using some other independent part of this database for the performance evaluation. However, cross-database performance evaluation (\textit{CrossDB}) is highly desirable to address limitations of currently available palmprint recognition methods in the literature. Therefore \textit{CrossDB} performance evaluation results are also presented in this report which uses the network that is trained on some part of publicly available database while the test performance are reported using other or independent publicly available database with the respective protocols which have been used in the literature (to ensure fairness in the performance comparison). It should be noted that for both \textit{WithinDB} and \textit{CrossDB} configurations, training set and test set are totally separated, i.e., none of the palmprint images are overlapping between training set and the test set. Since our focus is more on extensive \textit{CrossDB} performance evaluation, we incorporated the largest subjects database from 600 different subjects for this task as detailed in the next section.
\vspace{-0.8cm}
\setlength{\tabcolsep}{5pt}
\begin{table}
  \setlength{\abovecaptionskip}{0pt}
  \setlength{\belowcaptionskip}{1pt}
\begin{center}
\caption{Summary of accuracy (average rank-one recognition rate) and equal error rate (EER) on three different contactless palmprint databases.}
\label{tab:experiment}
\begin{tabular}{|c|c|c|c|c|c|}
  \hline
  \multirow{2}{*}{ }    & \multicolumn{2}{c|}{IITD} & \multicolumn{2}{c|}{PolyU-IITD (600 subject)} & CASIA \\ \cline{2-6} 
  & Accuracy(\%) & EER(\%) & Accuracy(\%) & EER(\%) & EER(\%) \\ \hline
  DoN(TPAMI16) &99.15 & 0.68 & 98.3 & 0.329 & 0.53 \\ \hline
  RLOC & 99.00& 0.88 & 98.45 & 0.557 & 1.0 \\ \hline
  Competitive Code & 99.85& 1.0 & 98.45 & 0.435 & 0.76 \\ \hline
  Ordinal Code & 98.92& 1.25 & 98.48& 0.451 & 0.79 \\ \hline
  Ours-CrossDB & / & / & 98.6 & 0.267 & \textbf{0.51} \\ \hline
  Ours-WithinDB & \textbf{99.20}& \textbf{0.60}& \textbf{98.7} & \textbf{0.153} & / \\ \hline
\end{tabular}
\end{center}
\vspace{-0.8cm}
\end{table}

During our \textit{CrossDB} performance evaluation, all of the test configurations uses the IITD Left \cite{iitd_url} (all left hand palmprint images in this dataset) as the training set. The trained model is used for the performance evaluation using IITD Right (all the right hand palmprint images) which indicates \textit{WithinDB} performance. During the \textit{WithinDB} configuration, we used the left palmprint images for the training set and the right palmprint images for test or performance evaluation as it allows us to perform fair comparison, with the respective results from more recent approach DoN in TPAMI16 \cite{zheng20163d} which has shown outperforming results over several state of art methods. The same trained model which is trained using IITD left hand palmprint images is used for the Cross-DB performance evaluation using the largest subjects database made available from \cite{polyu_iitd_url} and also using the CASIA contactless palmprint image database from \cite{casia_url}. Thus the \textit{CrossDB} performance evaluation can illustrate the generalization capability of the proposed model when few or the training samples from other databases are incorporated. The \textit{WithinDB} performance evaluation using \cite{polyu_iitd_url}, in addition to results from IITD \cite{iitd_url}, is also presented for comparative performance evaluation.
\begin{figure}[!t]
  \setlength\tabcolsep{0pt}
  \setlength{\abovecaptionskip}{1pt}
  \setlength{\belowcaptionskip}{-12pt}
  {\renewcommand{\arraystretch}{1}
  \begin{tabular}{ccc}
  \centering
  \includegraphics[width=0.33\textwidth]{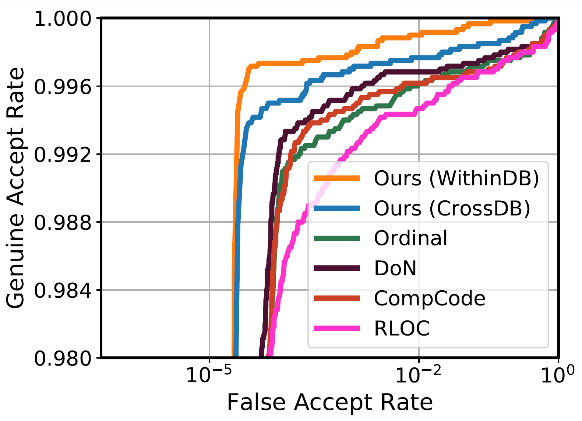}&\includegraphics[width=0.33\textwidth]{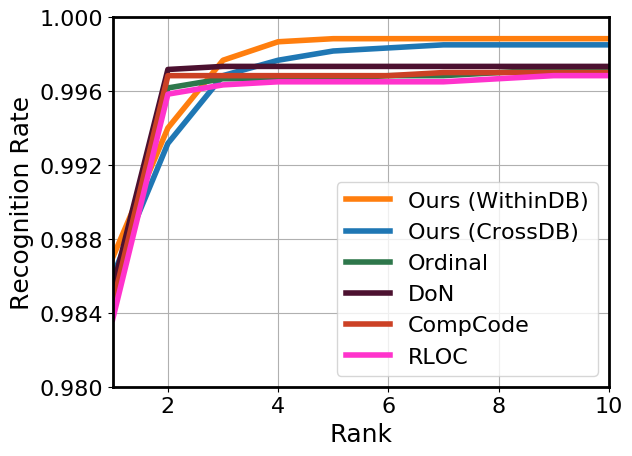}  & \includegraphics[width=0.33\textwidth]{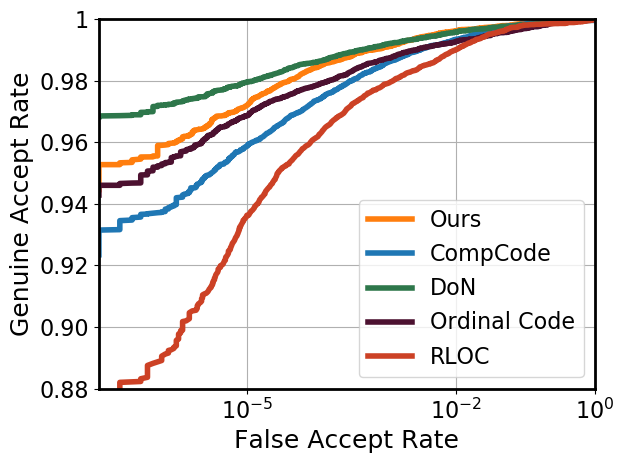} \\
  \scriptsize{(a)}&\scriptsize{(b)} & \scriptsize{(c)}
  \end{tabular}
  }
  \caption{(a) Comparative ROC and (b) corresponding CMC for \textit{WithinDB} and \textit{CrossDB} tests (600 subjects). (c) The ROC for \textit{CrossDB} evaluation for CASIA palmprint database \cite{casia_url} and other methods using respective/best parameters.}
  \label{fig:crossdb}  
\end{figure}
 \vspace{-0.7cm}
\subsection{Databases and Protocols}
\noindent\textbf{IITD Palmprint Database}. The IITD touchless palmprint database \cite{iitd_url} provides contactless palmprint images from the right and left hands of 230 subjects. There are 5 samples for each right hand or left hand. This database also provides $150\times 150$ pixels segmented palmprint images. In our experiments, the left hand palmprint images are used to train our model detailed in section 2 and all the 1300 right hand palmprint images are used for the performance evaluation. This protocol for test performance evaluation is exactly the same as in \cite{zheng20163d} and results in 1150 genuine matches and 263,350 imposter matches. The comparative performance using ROC, CMC (Figure \ref{fig:iitd}) and EER (Table \ref{tab:experiment}) is presented to ascertain the performance. The ROC, EER and the average rank-one recognition rate achieved from our approach indicates outperforming results.
\vspace{-0.1cm}
\subsection{Cross-Database Performance Evaluation}
Our \textit{CrossDB} performance evaluation were firstly focused on more recent database made available from [22] as this contactless palmprint database \cite{polyu_iitd_url} is acquired from 600 individuals which is the largest in the best of our knowledge. In our experiments, all the 6,000 palmprint images from the left hands were used for the test performance evaluation and the protocol is exactly the same as used for Figure \ref{fig:iitd} or the protocol used in \cite{zheng20163d}. Therefore, the test set for this \textit{CrossDB} performance generated 6,000 genuine and 3,594,000 imposter matches. Figure \ref{fig:crossdb} illustrates comparative ROC, CMC and respective EER is presented in Table \ref{tab:experiment}. This figure also illustrates \textit{WithinDB} performance which is achieved by training our model using other or all the right hand images for the same database. The results in Figure \ref{fig:iitd} (a)-(b) indicates that our model can achieve outperforming results and the performance is further improved for the \textit{WithinDB} case or when the model trained from the right hand images for the same database is used for the performance evaluation.

Another contactless palmprint database available in public domain is from \cite{casia_url}. This CASIA palmprint database contains 5239 palmprint images from 301 individuals. We also employed this database for the \textit{CrossDB} performance evaluation and used the model trained on IITD database (same as for results in Figure \ref{fig:iitd} or \textit{CrossDB} in Figure \ref{fig:crossdb}) for the performance evaluation. Our all experiments on this CASIA database used the same matching protocol as used in \cite{zheng20163d} to ensure fairness in the comparison. Therefore as in \cite{zheng20163d}, we also generated 13,692,466 match scores, which consisted of 20,567 genuine and 13,689,899 imposter match scores. Figure \ref{fig:crossdb} (c) illustrates comparative ROC performance and Table \ref{tab:experiment} provides respective EER from the \textit{CrossDB} performance. It is worth to underline that comparison is here with the same result as in \cite{zheng20163d}, which uses heavy tuning of parameters while our results are on unseen or \textit{CrossDB} evaluation protocol. Due to small number of images (only three) per subject, \textit{WithinDB} evaluation was not performed and is of least interest. It can be observed from these results that in terms of EER our model performs better than best of results in \cite{zheng20163d} while the performance from Figure \ref{fig:crossdb} is otherwise but quite competing.

\subsection{Discussion}
We also performed comparative performance evaluation from our method against other popular deep learning architectures that are widely used for various recognition tasks. The details on the such configurations considered for the performance evaluation is provided in the following.
\begin{itemize}
  \item \textbf{\textit{CNN}}+\textbf{\textit{Triplet Loss}} \par
  Pre-trained CNN based methods are most widely employed in the deep learning configurations for the recognition tasks \cite{Schroff2015FaceNet,szegedy2015going} and therefore also be interesting and worth evaluating. We select VGG-16 as our baseline test architecture which has achieved superior performance for many recognition problem and is widely used in other tasks. We replace the last fully connected class layer with another fully connected feature layer for matching the features. We froze the basic feature extraction layers in VGG-16 during the training phase and just fine-tuned the newly added fully connected layer using the given training dataset, i.e. IITD Left palmprint images in our experiment.
  \item \textbf{\textit{Fully convolutional network}}+\textbf{\textit{Extended triplet loss}} \par
  The fully convolutional network (FCN) was originally developed for the semantic segmentation \cite{long2015fully}. Recently \cite{zhao2017towards} combines FCN and extended triplet loss (ETL) to achieve the state-of-art performance for the iris recognition task. Since this work also employs bit-shifting in the original triplet loss function, it is important to comparatively ascertain the performance from our model over this method.
  \item \textbf{\textit{Residual feature network}}(\textbf{\textit{RFN}})+\textbf{\textit{Triplet loss}} \par
  Comparative evaluation has also been performed using the RFN used in our model and the original triplet loss function instead of the soft shifted triplet loss introduced in section 2.3. Such comparison is performed to ascertain the merit of SSTL for the problem considered in this report.
  \begin{figure}
  \setlength{\abovecaptionskip}{1pt}
  \setlength{\belowcaptionskip}{-12pt}
  \begin{center}
  \includegraphics[width=2.0in]{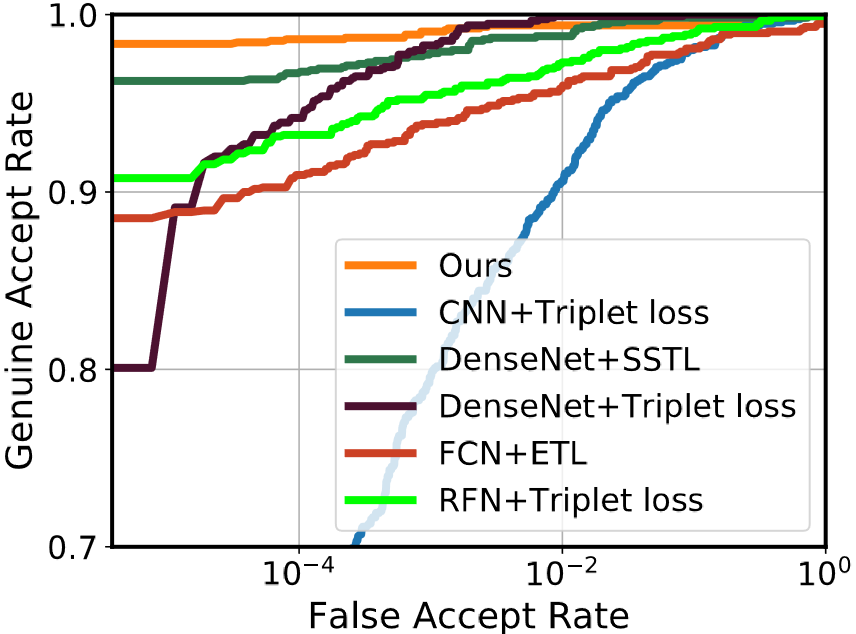}
  \caption{The ROC curves for typical deep learning architecture in the literature using contactless palmprint database in \cite{iitd_url}.}
  \label{fig:discuss}
  \end{center}
\end{figure}
  \item \textbf{\textit{DenseNet}}+\textbf{\textit{Soft shifted triplet loss/triplet loss}} \par
  We also compared our method against a very popular deep learning architecture, densely connected convolutional network (DenseNet) which has shown to offer significant performance improvement over the state-of-the-art on many/most recognition tasks. In our experiments on palmprint image datasets, we use a basic DenseNet-BC structure with three dense blocks on $128\times 128$ input images and replace the last fully connected layer with one $1\times 1$ convolutional layer to perform SSTL. The initial convolution layer uses $5\times 5$ convolutions with stride of two.
\end{itemize}

The comparison with the other deep learning based methods was performed on IITD dataset, which we employed for \textit{WithinDB} configuration with the same protocol as in Figure \ref{fig:iitd} or in \cite{zheng20163d}. All above discussed models were trained on the IITD Left palmprint images and evaluated using the IITD Right palmprint images. The test set generated 1150 genuine match scores and 263,350 imposter match scores which were consistent for the comparisons. The hyper-parameters for the training processes have been carefully investigated to achieve best performance. Comparative performances using ROC is presented in Figure \ref{fig:discuss} while comparative storage and matching complexity for these methods is summarized in Table \ref{tab:complexity}.

It can be observed from Figure \ref{fig:discuss} that our newly developed architecture together with newly developed soft shifted loss outperforms other deep learning configurations. The CNN based configurations suggest that global and high level features extracted by CNN may not be suitable for the palmprint recognition problem. Relatively poor performance from (RFN + Triplet) illustrates that a soft matching term introduced by SSTL offers great benefit in addressing inherent variations in the feature map for the contactless palmprint identification. Comparative performance between (DenseNet + SSTL) and (DenseNet + Triplet) also supports such observation.

In all of our experiments, we train our network using Stochastic Gradient Descent (SGD) with standard backprop \cite{lecun1989backpropagation, rumelhart1986learning} and Adam \cite{kingma2014adam}. We start with a learning rate of 0.001 and the models are randomly initialized. The pre-defined margin $\alpha$ is set to 0.2. The maximum vertical shift size $H_s$ and horizontal shifting size $W_s$ are both fixed as 5. More details on triplet selection, our network(RFN), along with additional results on \textit{CrossDB} performance are provided in \textit{supplementary file}.
\setlength{\tabcolsep}{7pt}
\begin{table}
  \setlength{\abovecaptionskip}{0pt}
  \setlength{\belowcaptionskip}{0pt}
\newcommand{\tabincell}[2]{\begin{tabular}{@{}#1@{}}#2\end{tabular}}
\begin{center}
\caption{The Comparison of time and space complexity of different contactless palmprint matching methods (evaluated on Linux Ubuntu 14.04 x86\_64 with Quadro M6000 GPU under 10K average runs. The default shift size was set as 5 for \textit{SSTL}).}
\label{tab:complexity}
\begin{tabular}{|c|c|c|c|c|}
  \hline
  Approaches & \tabincell{c}{Number of \\ parameters} & \tabincell{c}{Feature \\ extraction} & Matching & Template size \\
  \hline
  CNN+Triplet loss & \~449M & 0.00745s & 0.00140s & 4096-d vector \\
  \hline
  DenseNet+SSTL & \~3.1M & 0.0235s & 0.049s & $32\times 32$ map \\
  \hline
  DenseNet+Triplet loss & \~3.1M & 0.0235s & 0.00040s & $32\times 32$ map \\
  \hline
  FCN+ETL & \~568K & 0.00142s & 0.0710s & $12\times 128$ map \\
  \hline
  RFN+Triplet loss & \~5.2M & 0.0062s & 0.00040s & $32\times 32$ map \\
  \hline
  \textbf{Proposed} & \textbf{\~5.2M} & \textbf{0.0062s} & \textbf{0.049s} & \textbf{$32\times 32$ map} \\
  \hline
\end{tabular}
\end{center}
\vspace{-0.8cm}
\end{table}

\section{Online Palmprint Identification}
In earlier sections, we discussed on our approach for the development of trained model to match contactless palmprint images. The performance evaluation presented in section 3 used automatically segmented contactless palmprint images in respective public databases. The success of this matcher for online palmprint identification also requires a deep learning based palm detector that can automatically detect palmprint or the region of interest from the presented hands. Therefore we also developed a palm detector that can detect palmprints under complex background which is also considered as a challenge using the conventional methods of palmprint detection available in the literature.
\vspace{-0.2cm}

\subsection{Palmprint Detection}
The online palmprint detector developed in our work is based on the Faster R-CNN introduced in \cite{Ren2017Faster}. It is composed of two modules. The first module is a deep fully convolutional network (CNN) that proposes possible object regions. The second module is a Fast R-CNN detector \cite{Girshick2015Fast} that uses these proposed regions to classify the palmprint ones. The entire system is a single, unified network for the palm detection, which employs the popular terminology of neural networks with "attention" mechanisms. The regional proposal network (RPN) modules in the system updates Fast R-CNN module on the specific regions of interest to detect palm region. Tensorflow based \cite{abadi2016tensorflow} implementation was incorporated for palmprint data and augmentation. Please see demo in \cite{YoutubeLink} for online palmprint detection under complex indoor and outdoor backgrounds.
\vspace{-0.3cm}
\begin{figure}
  \setlength{\abovecaptionskip}{0pt}
  \setlength{\belowcaptionskip}{-40pt}
  \begin{center}
  \includegraphics[width=4.8in]{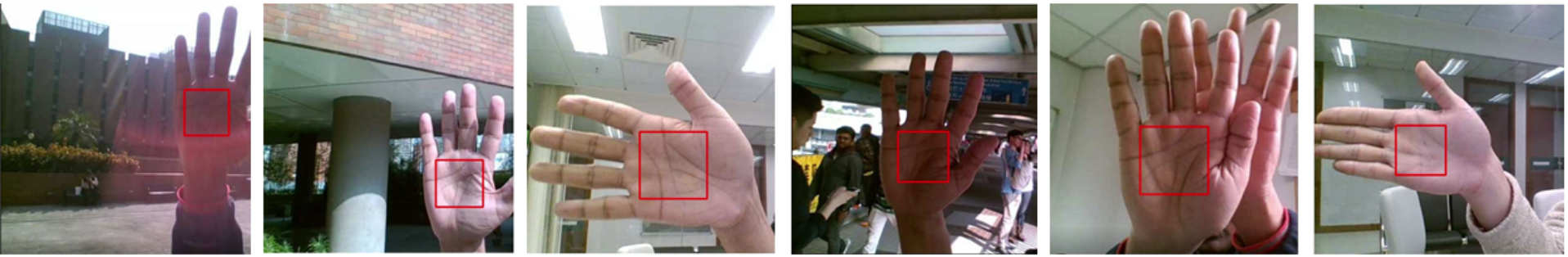}
  \caption{Sample images from our online system, running on a mobile laptop, depicting palmprint detection from hand images under complex backgrounds.}
  \label{fig:detect}
  \end{center}
\end{figure}
\subsection{Palmprint Dataset for Training and Detection}
We firstly acquired a set of videos under indoor and outdoor environment for developing the detector. These videos were acquired under 11 different environments with various postures and illuminations. The videos were then segmented at the interval of every 10 frames which resulted in a dataset of 3K raw palmprint images under varying backgrounds. This raw data is then augmented 10 times which resulted in a total of 30K palmprint images that were employed to train the palmprint detector.
\vspace{-0.4cm}
 \subsection{Performance Evaluation}
 We trained the palmprint detection model with 20K epochs which required about 7 hours for convergence on a single NVIDIA Quadro M6000. The test phase of the trained model requires an average of 0.101 seconds to generate the proposal bounding box with 300 RPN outputs.
 \vspace{-0.6cm}
 \setlength{\tabcolsep}{10pt}
 \begin{table}
   \setlength{\abovecaptionskip}{0pt}
   \setlength{\belowcaptionskip}{0pt}
 \begin{center}
 \caption{The mAP and recall value at different (IOU) threshold.}
 \label{tab:detector}
 \begin{tabular}{|c|c|c|c|c|c|c|}
   \hline
   \multirow{3}{*}{Experiments}    & \multicolumn{3}{c|}{mAP} & \multicolumn{3}{c|}{recall} \\ \cline{2-7} 
   & \multicolumn{3}{c|}{Overlap IOU threshold} & \multicolumn{3}{c|}{Overlap IOU threshold} \\ \cline{2-7}
   & 0.35 & 0.5 & 0.6 & 0.35 & 0.5 & 0.6 \\ \hline
   strategy(a) & 100.0 & 99.89 & 98.20 & 100.0 & 99.84 & 98.97 \\ \hline
   strategy(b) & 100.0 & 98.44 & 86.45 & 100.0 & 98.78 & 90.50 \\ \hline
 \end{tabular}
 \end{center}
 \vspace{-0.8cm}
 \end{table}
 
We also performed experiments to ascertain the performance during the test phase. These experiments are organized in two categories using the strategy and parameters: (a) Separate the dataset randomly in 0.9:0.1 ratio where 0.9 represents the fraction of data for training while and 0.1 represents remaining data for test/evaluation; (b) Separate the dataset by backgrounds, where 10 different background are mixed together to form the training data and the remaining background dataset is used for the test/evaluation. The first strategy tests randomly separate the dataset into 0.9:0.1 ratio while the second strategy selects one of different backgrounds as the test set to ascertain performance. Table \ref{tab:detector} shows the values obtained for mean average precision(mAP), and recall for the all experiments performed. One can observed that the network generates higher accuracy even up-to 0.5 to 0.6 overlap of IOU [36] threshold. Slight degradation in accuracy is observed when overlap IOU threshold is more than 0.6. The exact sample size for tests using strategy (a) and (b) is 3517 and 4770 respectively.
\vspace{-0.4cm}
\section{Conclusions and Further Work}
this report has developed a novel deep learning based contactless palmprint feature representation model, which can offer superior matching accuracy and high generalization capability for matching contactless palmprint images. We designed a soft-shifted triplet function to enable effective supervision in learning comprehensive and spatially corresponding residual features using fully convolutional network. Further extension of this work should focus on jointly evaluating the contactless palmprint detection and identification performance. Such evaluation requires new/public video dataset, from hands under complex backgrounds, from large number of subjects and is also part of our further work.

\bibliographystyle{splncs}
\bibliography{egbib}
\end{document}